\documentclass[acmsmall, authorversion]{acmart}

\usepackage{booktabs}

\AtBeginDocument{%
  \providecommand\BibTeX{{%
    \normalfont B\kern-0.5em{\scshape i\kern-0.25em b}\kern-0.8em\TeX}}}

\setcopyright{acmcopyright}
\copyrightyear{2020}
\acmYear{2020}
\acmDOI{10.1145/3385656}

\acmJournal{TKDD}
\acmVolume{14}
\acmNumber{3}
\acmArticle{34}
\acmMonth{5}

\begin{document}

\title{Better Classifier Calibration for Small Data Sets}

\author{Tuomo Alasalmi}
\email{tuomo.alasalmi@oulu.fi}
\orcid{0000-0001-9330-9995}
\author{Jaakko Suutala}
\email{jaakko.suutala@oulu.fi}
\orcid{0000-0002-6605-0057}
\author{Juha R{\"o}ning}
\email{juha.roning@oulu.fi}
\orcid{0000-0001-9993-8602}
\affiliation{%
  \institution{University of Oulu}
  \streetaddress{P.O. Box 4500}
  \postcode{90014}
  \city{Oulu}
  \country{Finland}
}
\author{Heli Koskim{\"a}ki}
\email{heli.koskimaki@ouraring.com}
\orcid{0000-0003-2760-2071}
\affiliation{%
  \institution{Oura Health Ltd.}
  \streetaddress{Elektroniikkatie 10}
  \city{Oulu}
  \country{Finland}}

%
%
%
%
%

\renewcommand{\shortauthors}{Alasalmi et al.}

\begin{abstract}
Classifier calibration does not always go hand in hand with the classifier's ability to separate the classes. There are applications where good classifier calibration, i.e. the ability to produce accurate probability estimates, is more important than class separation. When the amount of data for training is limited, the traditional approach to improve calibration starts to crumble. In this article we show how generating more data for calibration is able to improve calibration algorithm performance in many cases where a classifier is not naturally producing well-calibrated outputs and the traditional approach fails. The proposed approach adds computational cost but considering that the main use case is with small data sets this extra computational cost stays insignificant and is comparable to other methods in prediction time. From the tested classifiers the largest improvement was detected with the random forest and naive Bayes classifiers. Therefore, the proposed approach can be recommended at least for those classifiers when the amount of data available for training is limited and good calibration is essential.
\end{abstract}



\maketitle

\section{Introduction}
\label{intro}

In many machine learning applications, e.g. in the medical domain \cite{Connolly2017}, the models need to be explainable, or they will not be very useful. Obviously this means that the model needs to communicate to the user somehow what has led it to the given conclusion instead of just being a black-box \cite{Guidotti2018}. Another important factor in model explainability is the information how reliable the given prediction is. This property is called classifier calibration. A well calibrated classifier prediction is such that the predicted probability of an event is close to the proportion of the those events among a group of similar predictions \cite{Dawid1982}. However, the main design objective for classifiers tends to be good class separation and not accurate reliability estimation. Therefore, many classifiers are not well calibrated out of the box. To improve this probability estimate, accurate classifier calibration algorithms are needed. With accurate calibration, almost any model can output a good estimate of the probability that the decision it has made is indeed correct \cite{Niculescu2005}. Accurate probability estimates are also important for cost sensitive decision making \cite{zadrozny2001}.

\begin{figure}
    \centering
    \includegraphics[width=\linewidth]{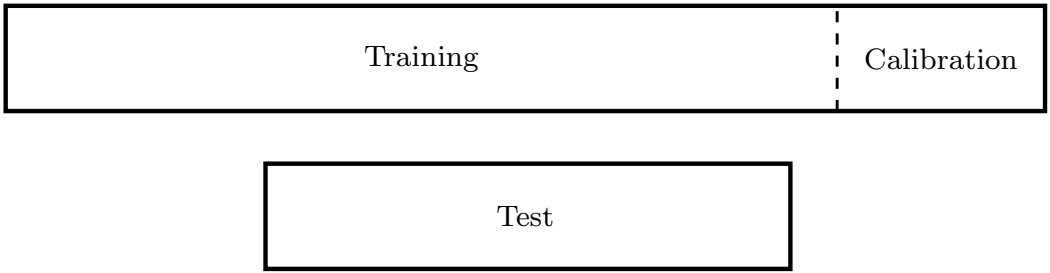}
    \Description{A figure showing how a data set is divided into training and test data sets. Calibration data set, if used, is split from the training data set.}
    \caption{Splitting of the available data into training, calibration, and test data sets. A part of the training data is reserved for calibration to avoid bias. Training data contains validation data for hyperparameter tuning.}
    \label{fig:data_splitting}
\end{figure}

For calibration algorithms to work well, a minimum of about 1000 to 2000 training samples are needed for the calibration data set depending on the learning algorithm to avoid overfitting. This is especially true for non-parametric calibration algorithms and calibration seems to improve further with even bigger calibration data sets \cite{Niculescu2005,NiculescuMizil2005}. To avoid biasing the calibration model, a separate calibration data set is needed. This means that the amount of training data in total needs to be large. E.g. if 10 \% of the training data set is used for calibration and the rest for modelling, a training data set with at least 10 000 samples is needed. In addition, a separate data set needs to be held out for testing. Figure \ref{fig:data_splitting} illustrates the data set partitioning. In many real world modelling tasks, however, relatively small data sets are quite common. As we will demonstrate in this article, traditional calibration algorithms fail to deliver on small data sets. But with our proposed data generation approach, calibration can often be improved despite the data set being small.

The rest of the article is structured as follows. Literature in calibration with a view on small data sets is briefly reviewed in Section \ref{sec:calibration}. In Section \ref{sec:experiments}, a set of experiments is described. The results of the experiments are summarized in Section \ref{sec:results} and presented in more detail in the Appendix. To conclude the article, the results are discussed in Section \ref{sec:discussion}.

\section{Classifier calibration}
\label{sec:calibration}


There are three main categories of calibration techniques. These are the parametric calibration algorithms such as Platt scaling \cite{Platt1999} and the non-parametric histogram binning \cite{Zadrozny2001binning} and isotonic regression \cite{Zadrozny2002} algorithms. In Platt scaling, a sigmoid function is fit to the prediction scores to transform prediction scores into probabilities. It was originally developed to improve calibration of support vector machines (SVM) and might not be the right transformation for many other classifiers. In binning, the prediction scores of a classifier are sorted and divided into bins of equal size. When we predict a test example, its prediction score can then be transformed into an estimated probability of belonging to a particular class by calculating the frequency of training samples belonging to that class in the corresponding bin. As drawbacks to binning, the number of bins needs to be specified and the probability estimates are discontinuous at bin boundaries. Also, depending on the classifier used, the prediction scores of classifier might not be uniformly distributed causing some bins to have significantly less, even zero, examples than others. Several methods have tried to overcome these problems, such as adaptive calibration of predictions (ACP) \cite{Jiang2012}, selection over Bayesian binnings (SBB) and averaging over Bayesian binnings (ABB) \cite{Naeini2015a}, as well as Bayesian binning into quantiles (BBQ) \cite{Naeini2015b}. In isotonic regression, a monotonically increasing function is used to map the prediction scores into probabilities. Isotonic regression is not continuous in general and can have undesirable jumps. To alleviate these problems, smoothing can be used \cite{Jiang2011}. In practice, however, the isotonicity assumption does not always hold \cite{Naeini2015a}. This makes isotonic regression sub optimal in these cases albeit quite effective \cite{Niculescu2005} regardless.


For the isotonicity constraint to hold true, the ranking imposed by the classifier would need to be perfect which is rarely true with real-world data sets. An ensemble of near-isotonic regression (ENIR) \cite{Naeini2018} allows violations of the ranking ordering and uses regularization to penalize the violations. In ENIR, a modified pool adjacent violators algorithm is used to find the solution path to a near isotonic regression problem \cite{Tibshirani2011} and Bayesian information criterion (BIC) scoring is used to combine the generated models. This ensemble is then used to post-process the classifier prediction scores to map them into calibrated probabilities. In their experiments, ENIR was on average the best performing calibration algorithm when compared to isotonic regression and BBQ with naive Bayes (NB), logistic regression, and SVM classifiers. Similarly, to what was accomplished with isotonic regression \cite{Zadrozny2002}, ENIR can be extended to multi-class problems whereas the Bayesian binning models cannot.

\subsection{Calibrating small data sets}
\label{sec:algorithm}

As already stated, to avoid biasing the calibration algorithm, a separate calibration data set is needed and it needs to be large enough to avoid overfitting. These constraints make the use of traditional calibration algorithms challenging with small data sets. For random forest (RF) classifiers, Out-of-Bag samples can be used so that the whole training data set can be utilized for both calibration and classifier training \cite{Bostrom2008}. An exact Bayesian model would not need calibration but as the true data distribution is not known in practice, we cannot construct such model. Instead, we can try to improve calibration by generating calibration data by Monte Carlo cross validation. The generation of calibration data can work, as we have previously shown, at least for isotonic regression calibration with the naive Bayes classifier \cite{Alasalmi2018}.

In our previous work \cite{Alasalmi2018}, two algorithms were suggested for calibration data generation. In the first stage, Monte Carlo cross validation is used to generate as many data points as desired. These value pairs consisting of the true class labels and the prediction scores can be used directly to tune the calibration algorithm. This is called the Data Generation (DG) model. The generated value pairs can be grouped and the average prediction scores along with the fraction of positive class labels in the group can be used for the calibration algorithm tuning. This model is called the Data Generation and Grouping (DGG) model. Detailed description of the process is not repeated here and the reader is instead referred to the original publication for details. In this work we will test the proposed data generation approach with the newer improved calibration algorithm ENIR and with more classifiers.

\section{Experiments}
\label{sec:experiments}
To test the effectiveness of using data generation for calibrating classifiers with small data sets, a set of experiments was set up. ENIR was used as the calibration algorithm because as a non parametric algorithm it should work equally well with all classifiers. In addition, Platt scaling was used with SVM. Representatives from top performing classifier groups were selected for the experiments and their calibration performance with different calibration scenarios was compared with two Bayesian classifiers.

To serve as control, we used the uncalibrated prediction scores of each classifier. This calibration scenario is referred in the results as Raw. In this case, as there was no need for a separate calibration data set, all data points in the training data set were used for classifier training. To test if the raw prediction scores could be improved by calibration to more closely resemble posterior probabilities, the calibration algorithm ENIR was used in four different settings. First, ENIR was used in the recommended way, i.e. a separate calibration data set was held out from the training data set that was not used for classifier training but only for tuning the calibration model. Size of the calibration data was set to 10 \% of the training data set and the remaining 90 \% was used for training the classifier. This scenario is called ENIR in the results. Second, ENIR was used like the algorithm's creators, i.e. the full training data set was used for both training the classifier and to tune the calibration model. This scenario is called ENIR full. The DG and DGG algorithms were also used with ENIR calibration. These are called DG + ENIR and DGG + ENIR, respectively. With the SVM classifier, Platt scaling was used with either a separate calibration data set as described above or with the full training data set. These are called Platt and Platt full in the results. Finally, the Out-of-Bag sample was used with ENIR calibration in the case of RF. This is called ENIR OOB in the results. R and Matlab code for carrying out the experiments is available on GitHub\footnote{\url{https://github.com/biovaan/Calibration}}.

There are literally hundreds of different classifiers available to use. Each of them has its place but not all of them perform equally well when compared over a diverse set of problems \cite{Cernadas2014}. For our experiments we chose a representative from each of the top performing classifier groups, namely a random forest, an SVM, and a feed forward neural network (NN) with a single hidden layer. In addition, a naive Bayes classifier was tested as it is computationally simple, easy to interpret, and surprisingly accurate despite of the often unrealistic assumption of feature independence. Also, the prediction scores of naive Bayes are not well calibrated which makes it a good candidate for this experiment \cite{Domingos1997}. In addition, two Bayesian classifiers were used that should produce well calibrated probabilities without separate calibration. These were Bayesian logistic regression (BLR) \cite{Gelman2008} which is a parametric linear classifier and Gaussian process classifier (GPC) \cite{Williams1998} which is non-parametric and nonlinear when nonlinear covariance function such as squared exponential is used. We tested the GPC implemented with expectation propagation (EP) approximation. Markov chain Monte Carlo (MCMC) sampling approximation of GPC can be considered the gold standard of GPC approximations but it is computationally very complex whereas EP approximation has been proven to have very good agreement with MCMC for both predictive probabilities and marginal likelihood estimates for fraction of the computational cost \cite{Kuss2005}.

RF was implemented using the R package randomForest. The default number of trees (500), $ntree$, was used and the hyperparameter $mtry$ was tuned by increments or decrements of two based on the Out-of-Bag error estimate. For SVM, the R package e1071 was used. A Gaussian kernel was used and the regularization parameter $cost$ was tuned with values $\{10^k\}^{11}_{-2}$. Good values for kernel spread hyperparameter $gamma$ were estimated based on the training data using the kernlab R package and the median value of the estimates was used \cite{caputo2002appearance}. The NN was implemented with the R package nnet. Hidden layer size was tuned in range from 1 to 9 neurons in increments of two and the hyperparameter $decay$ was tuned with values $\{10^k\}^0_{-4}$. As an activation function, a logistic function was used. For the Gaussian process classifier, GPML Matlab toolbox\footnote{\url{http://www.gaussianprocess.org/gpml/code/matlab/doc/}} implementation was used. A logistic likelihood function and a zero mean function was chosen and the covariance function was set to isotropic squared exponential covariance function which is in line with SVM with Gaussian kernel and regularization parameter $cost$. The hyperparameters for length-scale and signal magnitude were tuned by minimizing the negative log marginal likelihood (i.e., type II maximum likelihood approximation) on training data set. With the non-Bayesian methods, in every case except RF, which used Out-of-Bag error estimate, the tuning process was done using 10-fold cross validation on the training data excluding the calibration data. naive Bayes was implemented with the R package e1071. Bayesian logistic regression was implemented using the R package arm and default hyperparameter values (i.e., Cauchy prior with scale 2.5) were used and model was fitted by approximate expectation maximization algorithm on the training data set.

\subsection{Evaluating calibration performance}

Classifier calibration performance can be evaluated visually using a calibration plot or more objectively with some error metrics. With small data sets, the amount of data limits the usefulness of the calibration plot so they were not used for evaluating calibration performance in our experiments. Below we will introduce two error metrics that are commonly used to evaluate classifier calibration. These metrics are used to compare calibration performance of different calibration scenarios in our experiments.

Logarithmic loss (logloss) is an error metric that gives the biggest penalty for being both confident and wrong about a prediction. It is therefore a good metric to evaluate classifier calibration especially if cost sensitive decisions are made based on the classifier outcome. Logarithmic loss is defined in Equation (\ref{eq:logloss}). In the equation $N$ stands for the number of observations, $M$ stands for the number of class labels, $log$ is the natural logarithm, $y_{i,j}$ equals $1$ if observation $i$ belongs to class $j$, otherwise it is $0$, and $p_{i,j}$ stands for the predicted probability that observation $i$ belongs to class $j$. A smaller value of logarithmic loss means better calibration.

Mean squared error (MSE) is another metric that is often used to evaluate classifier calibration. The smaller the MSE value of a classifier, the better the calibration. However, MSE puts less emphasis on single confident but wrong decisions made by the classifier. It is defined in Equation (\ref{eq:mse}) where $N$ stands for the number of observations, $y_{i}$ is $1$ if observation $i$ belongs to the positive class, otherwise it is $0$, and $p_{i}$ is the predicted probability that observation $i$ belongs to the positive class. As with logloss, a smaller value of MSE means better calibration.

\begin{equation}
logloss = -\frac{1}{N} \sum_{i=1}^{N} \sum_{j=1}^{M} y_{i,j} log(p_{i,j})
\label{eq:logloss}
\end{equation}

\begin{equation}
MSE = \frac{\sum_{i=1}^{N} (y_{i} - p_{i})^2}{N}
\label{eq:mse}
\end{equation}

To test the performance of each approach to calibration with each of the classifiers, the following test sequence was ran. Features were standardized to have zero mean and unit variance and near zero variance features were deleted. Depending on the calibration scenario, the data set was divided into two or three parts as in Figure \ref{fig:data_splitting}. These were training and test data sets and in the ENIR and Platt scenarios, a separate calibration data set was split off from the training data set. In the Raw scenario, logloss and MSE were calculated on the raw prediction scores obtained with each classifier from the separate test data set. In the ENIR calibration scenario, the slightly smaller training data set was used to train each classifier and the prediction scores were calibrated using the ENIR algorithm that was tuned with the separate calibration data set. In ENIR full scenario, the whole training data set was used for both training the classifiers and tuning the ENIR algorithm. Finally the prediction scores from predicting the test data points were calibrated and the error metrics calculated. In DG + ENIR and DGG + ENIR scenarios, the corresponding algorithm was used to create a calibration data set that was then used to tune the ENIR algorithm. The whole training data set was used to train the classifiers and the test data set prediction scores were calibrated and error metrics calculated. Threshold used for classification was selected using the calibrated training data set so that the selected threshold maximized the classification rate. In addition to measuring the error metrics, each calibration scenario's computation time was also measured.

To be able to test the differences between calibration scenarios, a stratified 10-fold cross validation was used to create the data samples. A 5 $\times$ 2CV t-test \cite{Dietterich1998} or a combined 5 $\times$ 2CV F-test \cite{Alpaydm1999} has been suggested to be used to detect differences in classifier performance because of a lower Type I error. The lower Type I error, however, does not come without a compromise, namely higher Type II error (i.e. lower power). The lower power seems to be highlighted in our own experiments with small data sets as the inherent variance between the results on different folds is quite high. Therefore, cross validation was selected as the sampling method in our experiments and a Student's paired t-test with unequal variance assumption and the Welch modification to the degrees of freedom \cite{welch1947} was used to determine if there was a difference between calibration scenarios.

\subsection{Tests with synthetic data}

A synthetic data set, where true posterior probabilities can be calculated, was used to verify that the proposed data generation algorithms can indeed help improve calibration on small data sets. MSE and logloss are proper measures of calibration performance \cite{Kull2015} but in theory it is possible that with discrete labels even improvements in these calibration error metrics do not equate with more accurate probabilities. Instead, they could indicate that a higher probability was assigned to positive predictions and lower probability to negative predictions. However, this kind of change in the probabilities should increase logarithmic loss unless classification error approaches zero. With synthetic data, the predicted probabilities can be compared to true probabilities where any improvement in error metrics can only come from a real improvement in the predicted probabilities.

The data set was generated by sampling from normal distributions that represent the positive and negative classes, sampling 100 instances from each class. The true probabilities were calculated as the ratio of the probability density functions of the distributions at the sample coordinates. Derivative features were engineered from the original features and the original features were not given to the classifiers. This was done to make the problem harder to the models so that estimating the probabilities was not trivial. The R code that was used to create the synthetic data set is available in GitHub with the rest of the code.

\subsection{Tests with real data}

Table \ref{tab:data_sets} presents the properties of the real data sets that were used in the experiments. If the problem was not already a binary classification, it was converted into one. With QSAR biodegradation data set \cite{mansouri2013} (Biodegradation) the task is to predict if the chemicals are readily biodegradable or not based on molecular descriptors. In Blood Transfusion Service Center data set \cite{Yeh20095866} (Blood donation), whether previous blood donors donated blood again in March 2007 or not is predicted. Contraceptive Method Choice data set (Contraceptive) is a subset of the 1987 National Indonesia Contraceptive Prevalence Survey. The task here is to predict the choice of current contraceptive method. As a positive class a combination of classes short-term and long-term were used and the no-use class was used as the negative class. Letter Recognition data set (Letter) is a data set of predetermined image features for handwritten letter identification. A variation of the data set was created by reducing it down into a binary problem of two similar letters. The letter Q was selected as the positive class and the letter O as the negative class. In the Mammographic mass data set \cite{elter2007prediction} the prediction task is to discriminate benign and malignant Mammographic masses based on BI-RADS attributes and the patient's age. Malignant outcome served as the positive class and benign outcome as the negative class. The Titanic data set is from a Kaggle competition where the task is to predict whom of the passengers survived from the accident. Passenger name, ticket number, and cabin number were excluded from the features and only entries without missing values were used. All data sets used in the experiments are freely available from the UCI machine learning repository \cite{Dua2019} except the Titanic data set which is available from Kaggle.

\begin{table}
\centering
\caption{Data set properties.}
\label{tab:data_sets}       
\begin{tabular}{lcccc}
\toprule
Data set & Samples & Features & Positive class & Calibration samples \\
\midrule
Biodegradation & 1055 & 41 & 32 \%  & 94 \\
Blood donation & 748 & 4 & 24 \%  & 67 \\
Contraceptive & 1473 & 9 & 57 \%  & 132 \\
Letter & 1536 & 16 & 51 \%  & 138 \\
Mammographic mass & 831 & 4 & 48 \%  & 74 \\
Titanic & 714 & 7 & 41 \% & 64 \\
\bottomrule
\end{tabular}
\end{table}

\section{Results}
\label{sec:results}

The synthetic data set was used to verify that the proposed approach does indeed improve probability estimates and not just calibration error metrics with discrete labels. Mean squared errors with each classifier and calibration scenario are presented in Table \ref{tab:synthetic}. With the synthetic data, MSE was calculated using the true probabilities, not discrete labels.

\begin{table}
\centering
\caption{Mean squared error of different classifiers and calibration scenarios on the synthetic data set.}
\label{tab:synthetic}       
\begin{tabular}{lllllllll}
\toprule
Classifier & No Cal. & ENIR & E.full & DG & DGG & OOB & Platt & P.full \\
\midrule
NB & 0.072 & 0.129$^{*}$ & 0.082$^{*}$ & 0.071 & 0.072 &  &  &  \\
SVM & 0.039 & 0.096$^{*}$ & 0.064$^{*}$ & 0.040 & 0.039 &  & 0.074$^{*}$ & 0.053$^{*}$ \\
RF & 0.052 & 0.088$^{*}$ & 0.092$^{*}$ & 0.041$^{*}$ & 0.039$^{*}$ & 0.041 &  &  \\
NN & 0.047 & 0.106$^{*}$ & 0.053 & 0.039$^{*}$ & 0.039$^{*}$ &  &  &  \\
BLR & 0.063 &  &  &  &  &  &  &  \\
GPC & 0.041 &  &  &  &  &  &  &  \\
\bottomrule
\addlinespace[\belowrulesep]
\multicolumn{9}{p{0.8\linewidth}}{\footnotesize{Average results of 10-fold cross validation. Lower value of mean squared error indicates better calibration performance. $^{*}$ Significantly different from No Cal., $p < 0.05$. Significance of the difference determined with Student's paired t-test on 10-fold cross validation results.}}
\end{tabular}
\end{table}

\begin{figure}
    \centering
    \includegraphics[width=\linewidth]{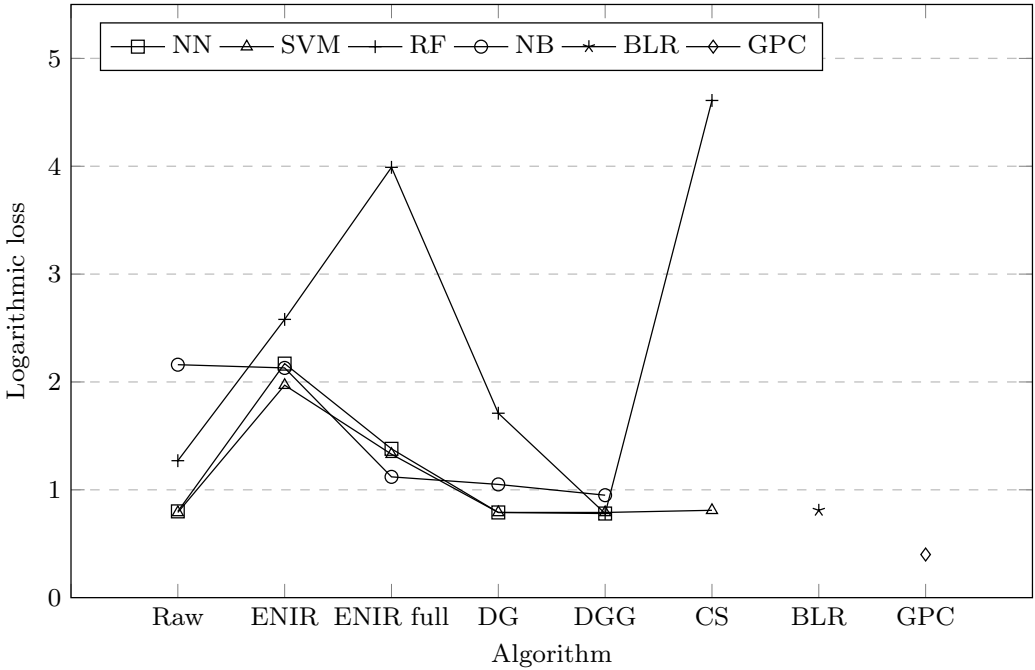}
    \Description{A line with markers chart showing the average logarithmic loss for different classifiers and calibration scenarios.}
    \caption{Average logarithmic loss for different classifiers and calibration scenarios. Classifier specific (CS) means Out-of-Bag samples with ENIR calibration for RF and Platt scaling with the full training set for SVM. Lower value of logloss indicates better calibration performance.}
    \label{fig:average_logloss}
\end{figure}
\begin{figure}
    \centering
    \includegraphics[width=\linewidth]{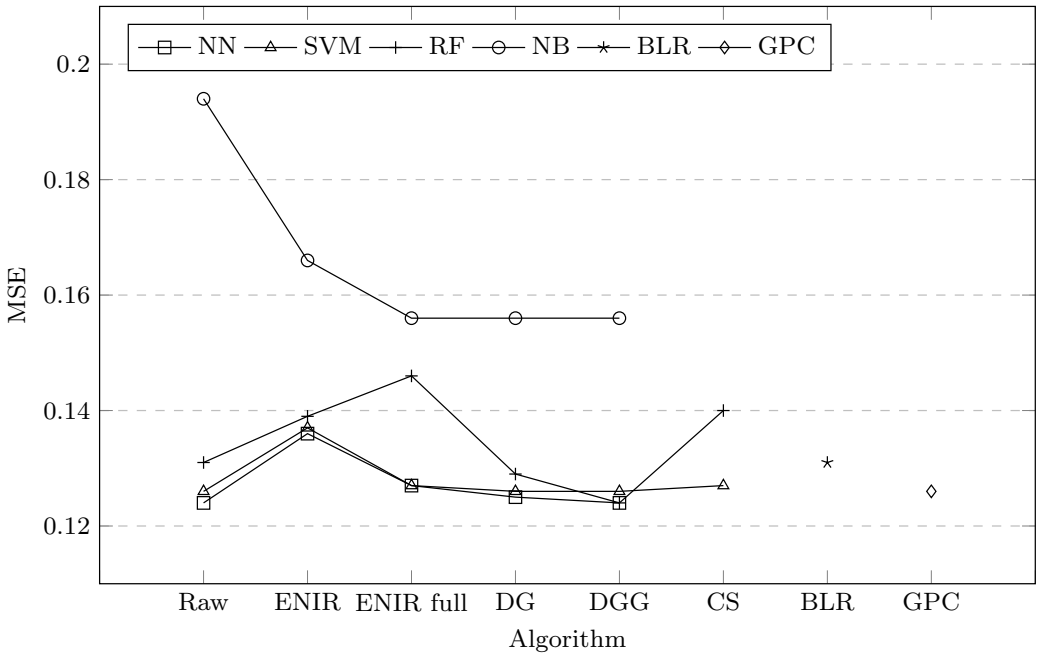}
    \Description{A line with markers chart showing the average mean squared error for different classifiers and calibration scenarios.}
    \caption{Average mean squared error for different classifiers and calibration scenarios. Classifier specific (CS) means Out-of-Bag samples with ENIR calibration for RF and Platt scaling with the full training set for SVM. Lower value of MSE indicates better calibration performance.}
    \label{fig:average_mse}
\end{figure}

Results of the experiments with real data sets are presented here summarized and the full results are attached as Appendix. The average logarithmic loss of each classifier and calibration scenario combination are depicted in Figure \ref{fig:average_logloss} and the average mean squared error in Figure \ref{fig:average_mse}. The training times of each classifier and calibration scenario were measured on a computational server (Intel Xeon E5-2650 v2 @ 2.60GHz, 196GB RAM) and the results are shown in Table \ref{tab:times}.

\begin{table}
\centering
\caption{Average computation times of the different classifiers and calibration scenarios in seconds on the experiment data sets. The times include hyperparameter tuning, training the classifier, and steps needed for calibration. Prediction times are excluded as they can be considered negligible.}
\label{tab:times}       
\begin{tabular}{lllllllll}
\toprule
Classifier & No Cal. & ENIR & E.full & DG & DGG & OOB & Platt & P.full \\
\midrule
NB & 0.01 & 0.06 & 0.02 & 4.06 & 3.87 &  &  &  \\
SVM & 8.13 & 6.40 & 8.14 & 10.38 & 10.21 &  & 6.34 & 8.14 \\
RF & 1.73 & 1.59 & 1.73 & 7.19 & 7.19 & 1.60 &  &  \\
NN & 172 & 154 & 172 & 174 & 174 &  &  &  \\
BLR & 0.17 &  &  &  &  &  &  &  \\
GPC & 281 &  &  &  &  &  &  &  \\
\bottomrule
\end{tabular}
\end{table}

\subsection{Interpretation of the results}

With the synthetic data set, it can be seen that using ENIR calibration with either a separate calibration data set or with the full training data set lead to poorer probability estimates than were achieved without calibration on all tested classifiers. On random forest and neural network, improvements in predicted probabilities were achieved by using either the DG or the DGG algorithm to generate the calibration data for ENIR calibration. The calibration error was in these cases lowered to approximately the same level as is achieved with the Gaussian process classifier. SVM achieved a comparable error level without calibration and no further improvement was achieved with the proposed calibration approach but the calibration error did not increase either. Using Platt scaling did increase the calibration error. Calibration error of naive Bayes was higher than the best of the pack and stayed intact with the proposed approach. This suggests that naive Bayes, due to the model's assumptions, was not flexible enough to catch the feature interactions in the data and therefore no improvement was achievable with calibration, even with DG or DGG data generation.

On the real data sets, with only one exception, the Biodegradation data set with naive Bayes classifier, using ENIR with a separate calibration data split off from the training data fails to improve calibration and actually makes the calibration worse although the differences are not statistically significant in every case. This obviously results from a very small calibration data set and these kind of results have also been noted in the literature before. This observation was the main motivation behind this work. When using the same data for training both the classifier and the ENIR calibration algorithm (ENIR full), we get mixed results. With the naive Bayes classifier, the calibration improves statistically significantly over uncalibrated control on four of the six data sets but the improvement does not reach statistical significance on the other two data sets. With the other three classifiers calibration tends to deteriorate compared to the uncalibrated control. The decrease in calibration performance is statistically significant on three data sets with SVM and RF, and four data sets with the NN classifier. What is interesting and supports our hypothesis is that with RF classifier ENIR full performs worse than ENIR with the tiny but separate calibration data which indicates overfitting. However, overfitting of the calibration model did not happen with the other classifiers.

Of the classifier specific calibration scenarios, Platt scaling performed equally well or insignificantly better with the small but separate calibration data set and the whole training data set as the calibration data on all but one data set on which using the full training data lowered logloss. Platt scaling did on average better than ENIR full, however, it could not improve calibration on any of the tested data sets over the uncalibrated control. The premise of using Out-of-Bag samples for calibration with RF was that the whole training data could be used for calibration without biasing the calibration model. Our results do not support that notion completely, at least on these small data sets. When Out-of-Bag samples were used to tune the ENIR calibration algorithm, calibration performance was worse than ENIR calibration with a separate calibration data set or using the full training data set on four and better on two of the tested data sets although one of the better performances was not statistically significant. What is more important, though, is that ENIR tuned with Out-of-Bag samples could improve calibration over the uncalibrated control on only one of the data sets. On those four misbehaving cases mentioned above the calibration significantly decreased instead.

Our DG algorithm coupled with ENIR was able to improve calibration over the uncalibrated control with the naive Bayes classifier on five of the tested data sets. SVM calibration improved slightly on two of the data sets with DG + ENIR and RF calibration performance decreased on three of the data sets while it improved on one. With NN, DG + ENIR calibration performance was not statistically significantly different from the uncalibrated control. It did, however, perform equally well or better than ENIR or ENIR full. DGG with ENIR calibration on the other hand improved calibration over uncalibrated control on all data sets with the naive Bayes classifier and on five out of six data sets with the RF classifier, although one of the improvements did not reach statistical significance. Calibration of SVM was improved on one of the data sets with DGG + ENIR and unaffected on the others. With NN, performance was improved with DGG + ENIR on one and decreased on one while being neutral on the other three data sets. DGG performed better than ENIR full on all data sets with all classifiers although the differences were not statistically significant in every case.

As a comparison, Bayesian logistic regression and Gaussian process classifiers were tested on the same data sets because these classifiers are supposed to be well calibrated without separate calibration. BLR calibration was better than the best non-Bayesian classifier with DGG + ENIR calibration on one of the data sets but worse on all other data sets although one of the differences was not statistically significant. Also, classification rate of BLR was slightly lower on average than on the other classifiers except NB, although the difference was not statistically significant. Logloss for GPC was lowest of all classifiers and calibration scenarios on five of the data sets by a clear margin but higher on one of the data sets. MSE, however, was higher on three and lower on one of the data sets than with the best of the calibrated non-Bayesian classifiers. This discrepancy indicates that a higher proportion of mistakes made by GPC were truly uncertain and high confidence predictions were more often correct with GPC than with the other classifiers. Thus, it could be said that GPC is not overconfident as classifiers calibrated with the ENIR algorithm. This is definitely an advantage in applications where good calibration is needed.

Using ENIR calibration with a separate calibration data set lead to a slightly lowered classification rate with three classifiers because the calibration data cannot be used for training the classifier making the training data set smaller. NN, SVM, and GPC had the highest classification rate on these data sets. A slightly lower classification rate was observed with RF and BLR classifiers. None of these small differences were, however, statistically significant. naive Bayes could not compete with the other classifiers in accuracy.

Training and calibration of naive Bayes, SVM, RF, and BLR took on average only seconds. NN and the EP approximation of GPC were clearly more computationally complex but still acceptable on these small data sets with training times of a few minutes.

\subsection{Effect of class imbalance}
\label{sec:class_imbalance}

To test how class imbalance problem affects the proposed data generation methods, another experiment was set up as follows. The Letter data set was used so that one of the classes on turn was downsampled to either 100, 50, or 25 samples resulting in six different data sets with the percentage of the positive class ranging from 3 \% to 12 \%. Classification rate was above the percentage of the larger class in every case so the classifiers can be considered to have worked reasonably well despite the class imbalance \cite{kuhn2013applied}. Same experiments were run on these data sets as before with the other data sets. The results of these experiments are shown in Tables \ref{tab:calibration_class_imbalance_mse} and \ref{tab:calibration_class_imbalance_logloss}. SVM and NN were well calibrated on these data sets without calibration so they are omitted from the tables. Using ENIR on a separate calibration data set or the full training data set did increase calibration error significantly as did Platt scaling on SVM when a separate calibration data set was used. Other methods had no significant effect on calibration performance on these two classifiers.

Class imbalance did not have a noticeable effect on the effectiveness of DG or DGG paired with ENIR calibration. With NB and RF classifiers the calibration of the raw scores were not optimal as can be seen from the difference compared to the Bayesian classifiers. Therefore DG and DGG with ENIR were able to improve their calibration. As was the case with more balanced data sets, DG with ENIR calibration lead to more overconfident probability estimates, i.e. low MSE but somewhat higher logloss, than DGG with ENIR calibration.

\begin{table}
    \centering
    \caption{Effect of class imbalance on MSE on the subsampled Letter data sets.}
    \begin{tabular}{lllllll}
        \toprule
        Classifier & OQ$_{100}$ & QO$_{100}$ & OQ$_{50}$ & QO$_{50}$ & OQ$_{25}$ & QO$_{25}$ \\
        \midrule
            NB Raw & 0.085 & 0.110 & 0.049 & 0.066 & 0.037 & 0.030 \\
            NB ENIR & 0.080 & 0.061$^{*}$ & 0.043 & 0.045$^{*}$ & 0.035 & 0.028 \\
            NB ENIR full & 0.068$^{*}$ & 0.050$^{*}$ & 0.040$^{*}$ & 0.033$^{*}$ & 0.025$^{*}$ & 0.024 \\
            NB DG + ENIR & 0.069$^{*}$ & 0.051$^{*}$ & 0.040$^{*}$ & 0.035$^{*}$ & 0.025$^{*}$ & 0.023 \\
            NB DGG + ENIR & 0.069$^{*}$ & 0.051$^{*}$ & 0.039$^{*}$ & 0.034$^{*}$ & 0.025$^{*}$ & 0.023 \\
        \midrule
            RF Raw & 0.021 & 0.021 & 0.015 & 0.017 & 0.012 & 0.014 \\
            RF ENIR & 0.022 & 0.021 & 0.013 & 0.026 & 0.017 & 0.022$^{*}$ \\
            RF ENIR full & 0.016$^{*}$ & 0.017$^{*}$ & 0.012$^{*}$ & 0.016 & 0.010 & 0.014 \\
            RF OOB & 0.017$^{*}$ & 0.015$^{*}$ & 0.014 & 0.015 & 0.010 & 0.012 \\
            RF DG + ENIR & 0.016$^{*}$ & 0.015$^{*}$ & 0.012$^{*}$ & 0.014 & 0.010$^{*}$ & 0.013 \\
            RF DGG + ENIR & 0.015$^{*}$ & 0.015$^{*}$ & 0.012$^{*}$ & 0.014$^{*}$ & 0.010$^{*}$ & 0.015 \\
        \midrule
        Bayesian logistic regression & 0.019 & 0.015 & 0.014 & 0.010 & 0.009 & 0.009 \\
        Gaussian process & 0.009 & 0.018 & 0.013 & 0.011 & 0.009 & 0.011 \\
        \bottomrule
        \addlinespace[\belowrulesep]
        \multicolumn{7}{p{0.8\linewidth}}{\footnotesize{Average results of 10-fold cross validation. Lower value of MSE indicates better calibration performance. $^{*}$ Significantly different from Raw, $p < 0.05$. Significance of the difference determined with Student's paired t-test on 10-fold cross validation results.}}
    \end{tabular}
    \label{tab:calibration_class_imbalance_mse}
\end{table}

\begin{table}
    \centering
    \caption{Effect of class imbalance on logloss on the subsampled Letter data sets.}
    \begin{tabular}{lllllll}
        \toprule
        Classifier & OQ$_{100}$ & QO$_{100}$ & OQ$_{50}$ & QO$_{50}$ & OQ$_{25}$ & QO$_{25}$ \\
        \midrule
            NB Raw & 0.954 & 0.924 & 0.629 & 0.452 & 0.585 & 0.244 \\
            NB ENIR & 1.626$^{*}$ & 1.550 & 1.375 & 1.495$^{*}$ & 1.510 & 0.674$^{*}$ \\
            NB ENIR full & 0.532$^{*}$ & 0.576$^{*}$ & 0.580 & 0.468 & 0.343 & 0.649 \\
            NB DG + ENIR & 0.473$^{*}$ & 0.384$^{*}$ & 0.439 & 0.332 & 0.192$^{*}$ & 0.195$^{*}$ \\
            NB DGG + ENIR & 0.474$^{*}$ & 0.383$^{*}$ & 0.357$^{*}$ & 0.327 & 0.193$^{*}$ & 0.193$^{*}$ \\
        \midrule
            RF Raw & 0.171 & 0.160 & 0.111 & 0.126 & 0.095 & 0.105 \\
            RF ENIR & 1.100$^{*}$ & 0.838 & 0.475 & 1.069$^{*}$ & 0.828$^{*}$ & 0.755$^{*}$ \\
            RF ENIR full & 0.537 & 0.598 & 0.309 & 0.396 & 0.387 & 0.629$^{*}$ \\
            RF OOB & 0.341 & 0.177 & 0.249 & 0.248 & 0.307 & 0.165 \\
            RF DG + ENIR & 0.321 & 0.453 & 0.166 & 0.316 & 0.307 & 0.237 \\
            RF DGG + ENIR & 0.183 & 0.115$^{*}$ & 0.083$^{*}$ & 0.171 & 0.080$^{*}$ & 0.117 \\
        \midrule
        Bayesian logistic regression & 0.132 & 0.120 & 0.118 & 0.079 & 0.082 & 0.078 \\
        Gaussian process & 0.038 & 0.079 & 0.054 & 0.044 & 0.040 & 0.048 \\
        \bottomrule
        \addlinespace[\belowrulesep]
        \multicolumn{7}{p{0.8\linewidth}}{\footnotesize{Average results of 10-fold cross validation. Lower value of logloss indicates better calibration performance. $^{*}$ Significantly different from Raw, $p < 0.05$. Significance of the difference determined with Student's paired t-test on 10-fold cross validation results.}}
    \end{tabular}
    \label{tab:calibration_class_imbalance_logloss}
\end{table}

\section{Discussion}
\label{sec:discussion}

The choice of a classifier depends on the problem at hand. Accuracy, computational complexity and memory requirements (e.g. wearable device vs. cloud server), and need for explainability are some properties that need to be taken into account when choosing a classifier. One aspect of explainability is classifier calibration, i.e. can the posterior probability estimates of the classifier be trusted. Bayesian methods such as Bayesian logistic regression and Gaussian process classifiers should be fairly well calibrated out of the box but may not be the most accurate on average when tested on a wide array of problems. The top performing classifier groups have been shown to be random forests, support vector machines, and neural network variations. Our results indicate that SVM and NN calibration on the tested small data sets is fairly good but sometimes it can be further improved by the DGG method coupled with ENIR calibration. RF on the other hand almost always benefits from the DGG method coupled with ENIR. Gaussian Process classifier held on to the premise of good calibration on most of the tested data sets but RF with DGG coupled with ENIR calibration produced probabilities whose average MSE over all the data sets was actually lower than with GPC. The same is not true for logloss which suggests that ENIR might produce overconfident probability estimates. This discrepancy between performance on MSE and logloss is more pronounced with the DG algorithm as it does not use label smoothing like implicitly DGG does. This leads to clearly overconfident probability estimates with the DG approach coupled with ENIR calibration. The proposed methods are not adversely affected by even severe class imbalance as demonstrated in the experiments. The improvements in calibration error metrics do indicate a real improvement in the quality of the predicted probabilities which was verified by the tests wit a synthetic data set where true probabilities are known.

A slight drawback in DGG is that the number of samples and the group size parameter need to be set. Also, the calibration data points generated with DG are not necessarily uniformly distributed meaning that with a fixed bin size the bin width in DGG can vary. This can potentially affect calibration resolution negatively with prediction scores that fall inside the widest bins. These cases are rare, however, otherwise the bins would be narrower. A possible drawback of Gaussian process classifier is that full GPCs unlike e.g. SVMs are not sparse out of the box but need additional approximation approaches. This needs to be considered when training classifiers on large-scale problems but might not pose a problem on small data sets.

On these small data sets ENIR on its own, either with a separate calibration data set or with the whole training data set, performs poorly on all classifiers except naive Bayes which is known for its poor calibration. Extra computation time from doing DGG is negligible in the case of small data sets where it is mostly needed and therefore its use is recommended when better calibration is essential. This is especially true with at least classifiers such as random forest and naive Bayes.

\begin{acks}
The authors would like to thank Infotech Oulu, Jenny and Antti Wihuri Foundation, Tauno T\"{o}nning Foundation, and Walter Ahlstr{\"o}m Foundation for financial support of this work.
\end{acks}

\bibliographystyle{ACM-Reference-Format}
\bibliography{small_data_calibration}

\appendix

\section*{Appendix}
\subsection*{Full results}

Full results of our experiments are presented in Tables \ref{tab:biodeg}-\ref{tab:titanic}. The results in the tables are averages and standard deviations of the results for each fold in 10-fold cross validation. The statistical tests to determine if the differences between calibration conditions are statistically significant were done with Student's paired t-tests with unequal variance assumption. Bayesian logistic regression and Gaussian process classifiers were compared to the best performing of the other classifiers based on logarithmic loss of that classifier after calibrating the classifier with ENIR using DGG generated calibration data. Table \ref{tab:abreviations} lists the abbreviations used in the result tables.

\begin{table}[h]
    \centering
    \begin{tabular}{ll}
        \toprule
        Abbreviation & Description \\
        \midrule
        BLR & Bayesian logistic regression \\
        CR & Classification rate \\
        DG & Data Generation algorithm \\
        DGG & Data Generation and Grouping algorithm \\
        ENIR & Ensemble of near isotonic regressions \\
        MSE & Mean squared error \\
        Logloss & Logarithmic loss \\
        NB & Naive Bayes \\
        NN & Neural network \\
        OOB & Out-of-Bag \\
        RF & Random forest \\
        SVM & Support vector machine \\
        \bottomrule
    \end{tabular}
    \caption{List of abbreviations used in the results.}
    \label{tab:abreviations}
\end{table}

\begin{table}[b]
\centering
\caption{Classification rate, mean squared error, and logarithmic loss of different classifiers and calibration scenarios on the Biodegradation data set.}
\label{tab:biodeg}       
\begin{tabular}{llll}
\toprule
Scenario & CR (\%) & MSE & Logloss  \\
\midrule
NB Raw & 83.69 $\pm$ 4.08 & 0.248 $\pm$ 0.043 & 5.970 $\pm$ 1.179 \\
NB ENIR & 83.31 $\pm$ 4.53 & 0.135$^{*}\dagger$ $\pm$ 0.023 & 2.261$^{*}\dagger$ $\pm$ 1.492 \\
NB ENIR full & 82.65 $\pm$ 3.68 & 0.127$^{*}$ $\pm$ 0.023 & 1.099$^{*}$ $\pm$ 0.444 \\
NB DG + ENIR & 83.78 $\pm$ 3.96 &  0.128$^{*}$ $\pm$ 0.025 & 1.046$^{*}$ $\pm$ 0.464 \\
NB DGG + ENIR & 83.69 $\pm$ 4.08 &  0.127$^{*}$ $\pm$ 0.023 & 0.808$^{*}$ $\pm$ 0.114  \\
\midrule
SVM Raw & 87.20 $\pm$ 3.57 & 0.101 $\pm$ 0.019 & 0.678 $\pm$ 0.112 \\
SVM ENIR & 85.31 $\pm$ 2.87 & 0.113$^{*}$ $\pm$ 0.022 & 1.843$^{*}$ $\pm$ 1.161 \\
SVM ENIR full & 86.73 $\pm$ 3.93 & 0.105 $\pm$ 0.023 & 1.658$^{*}$ $\pm$ 0.759 \\
SVM Platt & 86.26 $\pm$ 3.83 & 0.106 $\pm$ 0.023 & 0.707$\dagger$ $\pm$ 0.119 \\
SVM Platt full & 87.20 $\pm$ 3.57 & 0.105 $\pm$ 0.025 & 0.761$^{*}\dagger$ $\pm$ 0.192 \\
SVM DG + ENIR & 86.82 $\pm$ 4.10 & 0.101$\dagger$ $\pm$ 0.020 & 0.673$^{*}\dagger\#$ $\pm$ 0.114 \\
SVM DGG + ENIR & 87.20 $\pm$ 3.57 & 0.101 $\pm$ 0.020 & 0.675$\dagger\#$ $\pm$ 0.113 \\
\midrule
RF Raw & 85.87 $\pm$ 4.67 & 0.097 $\pm$ 0.021 & 0.693 $\pm$ 0.159 \\
RF ENIR & 85.02 $\pm$ 3.75 & 0.114$^{*}$ $\pm$ 0.026 & 2.833$^{*}\dagger$ $\pm$ 1.502 \\
RF ENIR full & 85.87 $\pm$ 4.67 & 0.125$^{*}$ $\pm$ 0.039  & 6.667 $\pm$ 2.349 \\
RF ENIR OOB & 85.87 $\pm$ 4.96 & 0.097$\dagger$ $\pm$ 0.022 & 0.732$\dagger$ $\pm$ 0.152 \\
RF DG + ENIR & 85.59 $\pm$ 4.66 & 0.100$\dagger$ $\pm$ 0.024 & 1.046$^{*}\dagger$ $\pm$ 0.494 \\
RF DGG + ENIR & 86.82 $\pm$ 3.77 & 0.097$\dagger$ $\pm$ 0.023 & 0.688$\dagger$ $\pm$ 0.152 \\
\midrule
NN Raw & 84.83 $\pm$ 3.57 & 0.112 $\pm$ 0.027 & 0.848 $\pm$ 0.222 \\
NN ENIR & 85.12 $\pm$ 3.00 & 0.118 $\pm$ 0.020 & 1.875$^{*}$ $\pm$ 1.057  \\
NN ENIR full & 84.55 $\pm$ 3.25 & 0.120$^{*}$ $\pm$ 0.028 & 2.165$^{*}$ $\pm$ 1.542  \\
NN DG + ENIR & 84.93 $\pm$ 4.20 & 0.109$\dagger$ $\pm$ 0.022 & 0.767$\dagger$ $\pm$ 0.172 \\
NN DGG + ENIR & 84.83 $\pm$ 3.89 & 0.108$\dagger$ $\pm$ 0.022 & 0.709$^{*}\dagger$ $\pm$ 0.113 \\
\midrule
BLR & 85.78 $\pm$ 3.99 & 0.106$\ddagger$ $\pm$ 0.024 & 0.699 $\pm$ 0.132 \\
Gaussian process & 86.54 $\pm$ 3.82 & 0.106$\ddagger$ $\pm$ 0.020 & 0.352$\ddagger$ $\pm$ 0.045 \\
\bottomrule
\addlinespace[\belowrulesep]
\multicolumn{4}{p{0.8\linewidth}}{\footnotesize{Average results of 10-fold cross validation $\pm$ standard deviation. Lower values of MSE and logloss indicate better calibration performance. $^{*}$ Significantly different from Raw, $p < 0.05$. $\dagger$ Significantly different from ENIR full, $p < 0.05$. $\#$ Significantly different from classifier specific calibration, $p < 0.05$. $\ddagger$ Significantly different from RF DGG + ENIR, $p < 0.05$. Significance of the difference determined with Student's paired t-test on 10-fold cross validation results.}}
\end{tabular}
\end{table}

\begin{table}
\centering
\caption{Classification rate, mean squared error, and logarithmic loss of different classifiers and calibration scenarios on the Blood donation data set.}
\label{tab:blood}       
\begin{tabular}{llll}
\toprule
Scenario & CR (\%) & MSE & Logloss  \\
\midrule
NB Raw & 76.07 $\pm$ 1.82 & 0.186 $\pm$ 0.028 & 1.440 $\pm$ 0.438 \\
NB ENIR & 76.07 $\pm$ 1.61 & 0.176 $\pm$ 0.021 & 2.555$^{*}\dagger$ $\pm$ 1.292  \\
NB ENIR full & 75.54 $\pm$ 2.90 & 0.168$^{*}$ $\pm$ 0.014 & 1.271 $\pm$ 0.399 \\
NB DG + ENIR & 76.34 $\pm$ 2.17 & 0.167$^{*}$ $\pm$ 0.013 & 1.270 $\pm$ 0.398 \\
NB DGG + ENIR & 76.60 $\pm$ 2.69 &  0.166$^{*}$ $\pm$ 0.012 &  1.011$^{*}$ $\pm$ 0.066 \\
\midrule
SVM Raw & 79.15 $\pm$ 3.04 & 0.163 $\pm$ 0.012 & 1.013 $\pm$ 0.059 \\
SVM ENIR & 77.81 $\pm$ 3.72 & 0.179 $\pm$ 0.028 & 3.309$^{*}$ $\pm$ 2.893  \\
SVM ENIR full & 78.07 $\pm$ 2.82 & 0.162 $\pm$ 0.018 & 1.935$^{*}$ $\pm$ 0.711  \\
SVM Platt & 78.87 $\pm$ 4.18 & 0.174 $\pm$ 0.017 & 1.089$\dagger$ $\pm$ 0.138 \\
SVM Platt full & 79.15 $\pm$ 3.04 & 0.162 $\pm$ 0.015 & 1.010$\dagger$ $\pm$ 0.076 \\
SVM DG + ENIR & 79.01 $\pm$ 3.03 & 0.160 $\pm$ 0.016 & 0.994$\dagger$ $\pm$ 0.077 \\
SVM DGG + ENIR & 79.01 $\pm$ 2.98 & 0.161 $\pm$ 0.015 & 0.997$\dagger$ $\pm$ 0.072 \\
\midrule
RF Raw & 76.60 $\pm$ 5.26 & 0.169 $\pm$ 0.023 & 2.794 $\pm$ 1.539 \\
RF ENIR & 75.26 $\pm$ 5.72 & 0.181 $\pm$ 0.023 & 3.616 $\pm$ 1.989 \\
RF ENIR full & 76.47 $\pm$ 4.34 & 0.191$^{*}$ $\pm$ 0.032 & 3.031 $\pm$ 1.997  \\
RF ENIR OOB & 77.40 $\pm$ 5.43 & 0.181$^{*}\dagger$ $\pm$ 0.028 & 5.971$^{*}\dagger$ $\pm$ 1.980 \\
RF DG + ENIR & 76.73 $\pm$ 5.16 & 0.168$\dagger$ $\pm$ 0.020 & 2.880 $\pm$ 1.739 \\
RF DGG + ENIR & 77.40 $\pm$ 5.43 & 0.161$^{*}\dagger$ $\pm$ 0.016 & 0.997$^{*}\dagger\#$ $\pm$ 0.083 \\
\midrule
NN Raw & 80.21 $\pm$ 2.96 & 0.148 $\pm$ 0.016 & 0.934 $\pm$ 0.081 \\
NN ENIR & 79.95 $\pm$ 2.81 & 0.169$^{*}\dagger$ $\pm$ 0.025  & 2.981$^{*}$ $\pm$ 2.677  \\
NN ENIR full & 80.21 $\pm$ 3.27 & 0.149 $\pm$ 0.018 & 1.518$^{*}$ $\pm$ 0.590 \\
NN DG + ENIR & 80.47 $\pm$ 3.22 & 0.149 $\pm$ 0.015 & 0.937$\dagger$ $\pm$ 0.071 \\
NN DGG + ENIR & 79.81 $\pm$ 2.87 & 0.148 $\pm$ 0.015 & 0.931$\dagger$ $\pm$ 0.074 \\
\midrule
BLR & 78.47 $\pm$ 3.65 & 0.155$\ddagger$ $\pm$ 0.013 & 0.956$\ddagger$ $\pm$ 0.066 \\
Gaussian process & 79.14 $\pm$ 2.43 & 0.152$\ddagger$ $\pm$ 0.015 & 0.473$\ddagger$ $\pm$ 0.036 \\
\bottomrule
\addlinespace[\belowrulesep]
\multicolumn{4}{p{0.8\linewidth}}{\footnotesize{Average results of 10-fold cross validation $\pm$ standard deviation. Lower values of MSE and logloss indicate better calibration performance. $^{*}$ Significantly different from Raw, $p < 0.05$. $\dagger$ Significantly different from ENIR full, $p < 0.05$. $\#$ Significantly different from classifier specific calibration, $p < 0.05$. $\ddagger$ Significantly different from NN DGG + ENIR, $p < 0.05$. Significance of the difference determined with Student's paired t-test on 10-fold cross validation results.}}
\end{tabular}
\end{table}

\begin{table}
\centering
\caption{Classification rate, mean squared error, and logarithmic loss of different classifiers and calibration scenarios on the Contraceptive use data set.}
\label{tab:contraceptive}       
\begin{tabular}{llll}
\toprule
Scenario & CR (\%) & MSE & Logloss  \\
\midrule
NB Raw & 63.00 $\pm$ 4.22 & 0.258 $\pm$ 0.028 & 1.802 $\pm$ 0.290 \\
NB ENIR & 64.02 $\pm$ 3.93 & 0.234$^{*}\dagger$ $\pm$ 0.020 & 1.973$\dagger$ $\pm$ 0.861  \\
NB ENIR full & 62.19 $\pm$ 3.88 & 0.225$^{*}$ $\pm$ 0.013 & 1.367$^{*}$ $\pm$ 0.290 \\
NB DG + ENIR & 62.59 $\pm$ 3.70 & 0.225$^{*}$ $\pm$ 0.013 & 1.286$^{*}$ $\pm$ 0.058 \\
NB DGG + ENIR & 63.00 $\pm$ 4.28 &  0.226$^{*}$ $\pm$ 0.013 &  1.287$^{*}$ $\pm$ 0.058 \\
\midrule
SVM Raw & 71.62 $\pm$ 2.95 & 0.195 $\pm$ 0.011 & 1.153 $\pm$ 0.050 \\
SVM ENIR & 70.60 $\pm$ 3.03 & 0.204$^{*}$ $\pm$ 0.013  & 1.924$^{*}$ $\pm$ 0.743  \\
SVM ENIR full & 70.81 $\pm$ 3.09 & 0.197 $\pm$ 0.014 & 1.469$^{*}$ $\pm$ 0.380 \\
SVM Platt & 71.76 $\pm$ 2.58 & 0.197 $\pm$ 0.009 & 1.162$\dagger$ $\pm$ 0.043 \\
SVM Platt full & 71.62 $\pm$ 2.95 & 0.196 $\pm$ 0.014 & 1.162$\dagger$ $\pm$ 0.063 \\
SVM DG + ENIR & 71.56 $\pm$ 3.46 & 0.194 $\pm$ 0.012 & 1.194 $\pm$ 0.164 \\
SVM DGG + ENIR & 71.35 $\pm$ 3.32 & 0.194$\dagger\#$ $\pm$ 0.012 & 1.147$\dagger$ $\pm$ 0.053 \\
\midrule
RF Raw & 70.06 $\pm$ 4.02 & 0.196 $\pm$ 0.014 & 1.228 $\pm$ 0.153 \\
RF ENIR & 70.67 $\pm$ 4.07 & 0.197$\dagger$ $\pm$ 0.014 & 1.533$^{*}\dagger$ $\pm$ 0.342 \\
RF ENIR full & 71.35 $\pm$ 4.36 & 0.228$^{*}$ $\pm$ 0.020 & 4.350$^{*}$ $\pm$ 1.314 \\
RF ENIR OOB & 70.07 $\pm$ 4.51 & 0.218$^{*}$ $\pm$ 0.019 & 6.019$^{*}\dagger$ $\pm$ 1.358 \\
RF DG + ENIR & 69.79 $\pm$ 3.18 & 0.198$\dagger$ $\pm$ 0.012 & 1.454$\dagger$ $\pm$ 0.445 \\
RF DGG + ENIR & 69.93 $\pm$ 3.98 & 0.191$^{*}\dagger$ $\pm$ 0.011 & 1.123$^{*}\dagger\#$ $\pm$ 0.056 \\
\midrule
NN Raw & 71.22 $\pm$ 2.70 & 0.189 $\pm$ 0.015 & 1.120 $\pm$ 0.070 \\
NN ENIR & 70.61 $\pm$ 2.56 & 0.200$^{*}\dagger$ $\pm$ 0.010  & 1.977$^{*}\dagger$ $\pm$ 9.49  \\
NN ENIR full & 70.94 $\pm$ 3.38 & 0.189 $\pm$ 0.014 & 1.251 $\pm$ 0.383  \\
NN DG + ENIR & 71.28 $\pm$ 2.76 & 0.190 $\pm$ 0.011 & 1.167 $\pm$ 0.140 \\
NN DGG + ENIR & 71.49 $\pm$ 2.69 & 0.191 $\pm$ 0.012 & 1.129 $\pm$ 0.053 \\
\midrule
BLR & 68.30 $\pm$ 3.80 & 0.210$\ddagger$ $\pm$ 0.011 & 1.216$\ddagger$ $\pm$ 0.049 \\
Gaussian process & 71.49 $\pm$ 3.61 & 0.192 $\pm$ 0.011 & 0.570$\ddagger$ $\pm$ 0.028 \\
\bottomrule
\addlinespace[\belowrulesep]
\multicolumn{4}{p{0.8\linewidth}}{\footnotesize{Average results of 10-fold cross validation $\pm$ standard deviation. Lower values of MSE and logloss indicate better calibration performance. $^{*}$ Significantly different from Raw, $p < 0.05$. $\dagger$ Significantly different from ENIR full, $p < 0.05$. $\#$ Significantly different from classifier specific calibration, $p < 0.05$. $\ddagger$ Significantly different from RF DGG + ENIR, $p < 0.05$. Significance of the difference determined with Student's paired t-test on 10-fold cross validation results.}}
\end{tabular}
\end{table}

\begin{table}
\centering
\caption{Classification rate, mean squared error, and logarithmic loss of different classifiers and calibration scenarios on the Letter recognition data set.}
\label{tab:letter}       
\begin{tabular}{llll}
\toprule
Scenario & CR (\%) & MSE & Logloss  \\
\midrule
NB Raw & 84.24 $\pm$ 2.98 & 0.135 $\pm$ 0.023 & 1.060 $\pm$ 0.190 \\
NB ENIR & 84.18 $\pm$ 3.00 & 0.109$^{*}$ $\pm$ 0.018 & 1.307$\dagger$ $\pm$ 0.701 \\
NB ENIR full & 83.66 $\pm$ 1.91 & 0.104$^{*}$ $\pm$ 0.011 & 0.720$^{*}$ $\pm$ 0.224 \\
NB DG + ENIR & 84.38 $\pm$ 2.38 & 0.104$^{*}$ $\pm$ 0.012 & 0.647$^{*}$ $\pm$ 0.057 \\
NB DGG + ENIR & 84.05 $\pm$ 2.41 & 0.104$^{*}$ $\pm$ 0.012 & 0.648$^{*}$ $\pm$ 0.056 \\
\midrule
SVM Raw & 99.28 $\pm$ 0.54 & 0.006 $\pm$ 0.005 & 0.049 $\pm$ 0.029 \\
SVM ENIR & 99.22 $\pm$ 0.91 & 0.008 $\pm$ 0.008 & 0.377$^{*}$ $\pm$ 0.448 \\
SVM ENIR full & 99.28 $\pm$ 0.54 & 0.007 $\pm$ 0.006 & 0.171 $\pm$ 0.394 \\
SVM Platt & 99.02 $\pm$ 1.06 & 0.007 $\pm$ 0.006 & 0.082$^{*}$ $\pm$ 0.046 \\
SVM Platt full & 99.28 $\pm$ 0.54 & 0.006 $\pm$ 0.005 & 0.054 $\pm$ 0.049 \\
SVM DG + ENIR & 99.15 $\pm$ 0.71 & 0.006 $\pm$ 0.005 & 0.043$^{*}\#$ $\pm$ 0.033 \\
SVM DGG + ENIR & 99.28 $\pm$ 0.54 & 0.006 $\pm$ 0.005 & 0.044$^{*}\#$ $\pm$ 0.032 \\
\midrule
RF Raw & 97.53 $\pm$ 1.30 & 0.024 $\pm$ 0.007 & 0.210 $\pm$ 0.045 \\
RF ENIR & 97.33 $\pm$ 1.18 & 0.018$^{*}$ $\pm$ 0.009 & 0.433 $\pm$ 0.551 \\
RF ENIR full & 97.53 $\pm$ 1.30 & 0.019$^{*}$ $\pm$ 0.019 & 0.567 $\pm$ 0.677  \\
RF ENIR OOB & 97.79 $\pm$ 1.38 & 0.014$^{*}$ $\pm$ 0.008 & 0.137 $\pm$ 0.135 \\
RF DG + ENIR & 97.92 $\pm$ 1.27 & 0.014$^{*}$ $\pm$ 0.008 & 0.134 $\pm$ 0.157 \\
RF DGG + ENIR & 97.98 $\pm$ 1.07 & 0.013$^{*}$ $\pm$ 0.007 & 0.094$^{*}$ $\pm$ 0.047 \\
\midrule
NN Raw & 98.96 $\pm$ 0.83 & 0.008 $\pm$ 0.006 & 0.057 $\pm$ 0.039 \\
NN ENIR & 98.70 $\pm$ 1.01 & 0.011 $\pm$ 0.008 & 0.438$^{*}$ $\pm$ 0.356 \\
NN ENIR full & 98.96 $\pm$ 0.83 & 0.009 $\pm$ 0.007 & 0.346$^{*}$ $\pm$ 0.404  \\
NN DG + ENIR & 99.02 $\pm$ 0.84 & 0.008 $\pm$ 0.006 & 0.053$\dagger$ $\pm$ 0.033 \\
NN DGG + ENIR & 98.96 $\pm$ 0.83 & 0.008 $\pm$ 0.006 & 0.054$\dagger$ $\pm$ 0.034 \\
\midrule
BLR & 95.90$\ddagger$ $\pm$ 1.84 & 0.028$\ddagger$ $\pm$ 0.012 & 0.193$\ddagger$ $\pm$ 0.063 \\
Gaussian process & 98.11 $\pm$ 1.14 & 0.023$\ddagger$ $\pm$ 0.006 & 0.107$\ddagger$ $\pm$ 0.015 \\
\bottomrule
\addlinespace[\belowrulesep]
\multicolumn{4}{p{0.8\linewidth}}{\footnotesize{Average results of 10-fold cross validation $\pm$ standard deviation. Lower values of MSE and logloss indicate better calibration performance. $^{*}$ Significantly different from Raw, $p < 0.05$. $\dagger$ Significantly different from ENIR full, $p < 0.05$. $\ddagger$ Significantly different from SVM DGG + ENIR, $p < 0.05$. Significance of the difference determined with Student's paired t-test on 10-fold cross validation results.}}
\end{tabular}
\end{table}

\begin{table}
\centering
\caption{Classification rate, mean squared error, and logarithmic loss of different classifiers and calibration scenarios on the Mammographic mass data set.}
\label{tab:mammo}       
\begin{tabular}{llll}
\toprule
Scenario & CR (\%) & MSE & Logloss  \\
\noalign{\smallskip}\hline\noalign{\smallskip}
NB Raw & 77.62 $\pm$ 4.71 & 0.169 $\pm$ 0.036 & 1.292 $\pm$ 0.288 \\
NB ENIR & 78.47 $\pm$ 4.19 & 0.165$\dagger$ $\pm$ 0.027 & 2.116$\dagger$ $\pm$ 0.866 \\
NB ENIR full & 77.98 $\pm$ 4.87 & 0.153$^{*}$ $\pm$ 0.026 & 1.105 $\pm$ 0.396 \\
NB DG + ENIR & 78.34 $\pm$ 4.28 & 0.154$^{*}$ $\pm$ 0.027 & 1.036$^{*}$ $\pm$ 0.339 \\
NB DGG + ENIR & 78.09 $\pm$ 4.83 & 0.153$^{*}$ $\pm$ 0.026 &  0.958$^{*}$ $\pm$ 0.130 \\
\midrule
SVM Raw & 80.14 $\pm$ 4.25 & 0.150 $\pm$ 0.025 & 0.944 $\pm$ 0.119 \\
SVM ENIR & 78.94 $\pm$ 4.45 & 0.165$^{*}\dagger$ $\pm$ 0.025 & 2.203$^{*}\dagger$ $\pm$ 1.139 \\
SVM ENIR full & 79.66 $\pm$ 4.22 & 0.150 $\pm$ 0.027 & 1.394 $\pm$ 0.713 \\
SVM Platt & 79.78 $\pm$ 5.14 & 0.152 $\pm$ 0.024 & 0.964 $\pm$ 0.121 \\
SVM Platt full & 80.14 $\pm$ 4.25 & 0.151 $\pm$ 0.026 & 0.949 $\pm$ 0.127 \\
SVM DG + ENIR & 80.14 $\pm$ 3.82 & 0.152 $\pm$ 0.026 & 0.947 $\pm$ 0.128 \\
SVM DGG + ENIR & 80.14 $\pm$ 4.25 & 0.152 $\pm$ 0.026 & 0.951 $\pm$ 0.127 \\
\midrule
RF Raw & 80.50 $\pm$ 4.41 & 0.160 $\pm$ 0.030 & 1.556 $\pm$ 0.575 \\
RF ENIR & 80.02 $\pm$ 3.83 & 0.165 $\pm$ 0.029 & 3.729$^{*}\dagger$ $\pm$ 1.721 \\
RF ENIR full & 80.50 $\pm$ 4.04 & 0.159 $\pm$ 0.033 & 5.219$^{*}$ $\pm$ 1.757 \\
RF ENIR OOB & 80.50 $\pm$ 4.31 & 0.181$^{*}\dagger$ $\pm$ 0.040 & 10.12$^{*}\dagger$ $\pm$ 2.450 \\
RF DG + ENIR & 80.14 $\pm$ 4.25 & 0.157 $\pm$ 0.025 & 2.697$^{*}\dagger$ $\pm$ 0.819 \\
RF DGG + ENIR & 80.63 $\pm$ 4.31 & 0.148$^{*}\dagger$ $\pm$ 0.025 & 0.935$^{*}\dagger\#$ $\pm$ 0.115 \\
\midrule
NN Raw & 80.02 $\pm$ 4.68 & 0.148 $\pm$ 0.027 & 0.919 $\pm$ 0.139 \\
NN ENIR & 79.54 $\pm$ 4.38 & 0.156 $\pm$ 0.028  & 2.369$^{*}\dagger$ $\pm$ 1.320 \\
NN ENIR full & 79.78 $\pm$ 5.05 & 0.151 $\pm$ 0.030 & 1.384 $\pm$ 0.878 \\
NN DG + ENIR & 79.66 $\pm$ 5.30 & 0.151 $\pm$ 0.028 & 0.942 $\pm$ 0.143 \\
NN DGG + ENIR & 80.02 $\pm$ 4.68 & 0.151 $\pm$ 0.028 & 0.938$^{*}$ $\pm$ 0.141 \\
\midrule
BLR & 80.02 $\pm$ 4.74 & 0.145 $\pm$ 0.025 & 0.904$\ddagger$ $\pm$ 0.122 \\
Gaussian process & 81.10 $\pm$ 4.65 & 0.145$\ddagger$ $\pm$ 0.025 & 0.452$\ddagger$ $\pm$ 0.062 \\
\bottomrule
\addlinespace[\belowrulesep]
\multicolumn{4}{p{0.8\linewidth}}{\footnotesize{Average results of 10-fold cross validation $\pm$ standard deviation. Lower values of MSE and logloss indicate better calibration performance. $^{*}$ Significantly different from Raw, $p < 0.05$. $\dagger$ Significantly different from ENIR full, $p < 0.05$. $\#$ Significantly different from classifier specific calibration, $p < 0.05$. $\ddagger$ Significantly different from RF DGG + ENIR, $p < 0.05$. Significance of the difference determined with Student's paired t-test on 10-fold cross validation results.}}
\end{tabular}
\end{table}

\begin{table}
\centering
\caption{Classification rate, mean squared error, and logarithmic loss of different classifiers and calibration scenarios on the Titanic data set.}
\label{tab:titanic}       
\begin{tabular}{llll}
\toprule
Scenario & CR (\%) & MSE & Logloss  \\
\noalign{\smallskip}\hline\noalign{\smallskip}
NB Raw & 78.14 $\pm$ 4.51 & 0.170 $\pm$ 0.029 & 1.390 $\pm$ 0.430 \\
NB ENIR & 76.88 $\pm$ 4.01 & 0.176$\dagger$ $\pm$ 0.030 & 2.559$^{*}\dagger$ $\pm$ 1.592 \\
NB ENIR full & 77.58 $\pm$ 3.88 & 0.160$^{*}$ $\pm$ 0.024 & 1.163 $\pm$ 0.371 \\
NB DG + ENIR & 77.31 $\pm$ 3.60 & 0.160$^{*}$ $\pm$ 0.024 & 0.989$^{*}$ $\pm$ 0.116 \\
NB DGG + ENIR & 77.58 $\pm$ 4.33 & 0.160$^{*}$ $\pm$ 0.024 & 0.993$^{*}$ $\pm$ 0.114 \\
\midrule
SVM Raw & 81.65 $\pm$ 2.86 & 0.140 $\pm$ 0.015 & 0.900 $\pm$ 0.072 \\
SVM ENIR & 80.52 $\pm$ 3.32 & 0.153$^{*}\dagger$ $\pm$ 0.024 & 2.194 $\pm$ 1.903 \\
SVM ENIR full & 80.11 $\pm$ 3.76 & 0.140 $\pm$ 0.019 & 1.337 $\pm$ 0.599 \\
SVM Platt & 82.35 $\pm$ 3.18 & 0.144 $\pm$ 0.019 & 0.928 $\pm$ 0.103 \\
SVM Platt full & 81.65 $\pm$ 2.86 & 0.140 $\pm$ 0.018 & 0.899 $\pm$ 0.091 \\
SVM DG + ENIR & 81.93 $\pm$ 2.75 & 0.142 $\pm$ 0.013 & 0.907 $\pm$ 0.065 \\
SVM DGG + ENIR & 82.07 $\pm$ 2.80 & 0.141 $\pm$ 0.014 & 0.905 $\pm$ 0.067 \\
\midrule
RF Raw & 80.11 $\pm$ 4.30 & 0.139 $\pm$ 0.023 & 1.130 $\pm$ 0.454 \\
RF ENIR & 78.16 $\pm$ 3.97 & 0.158$^{*}$ $\pm$ 0.024 & 3.344$^{*}$ $\pm$ 1.819 \\
RF ENIR full & 80.81 $\pm$ 4.45 & 0.152$^{*}$ $\pm$ 0.029 & 4.090$^{*}$ $\pm$ 2.268 \\
RF ENIR OOB & 80.11 $\pm$ 3.71 & 0.149$^{*}$ $\pm$ 0.026 & 4.653$^{*}$ $\pm$ 2.869 \\
RF DG + ENIR & 81.09 $\pm$ 2.99 & 0.138$\#$ $\pm$ 0.017 & 2.019$^{*}\dagger\#$ $\pm$ 1.112 \\
RF DGG + ENIR & 80.24 $\pm$ 4.25 & 0.135$\dagger\#$ $\pm$ 0.017 & 0.863$\dagger\#$ $\pm$ 0.100 \\
\midrule
NN Raw & 80.24 $\pm$ 4.64 & 0.141 $\pm$ 0.022 & 0.903 $\pm$ 0.132 \\
NN ENIR & 78.98 $\pm$ 1.85 & 0.163$^{*}\dagger$ $\pm$ 0.019 & 3.375$^{*}\dagger$ $\pm$ 1.866 \\
NN ENIR full & 80.11 $\pm$ 4.39 & 0.144 $\pm$ 0.025 & 1.605$^{*}$ $\pm$ 0.684 \\
NN DG + ENIR & 80.25 $\pm$ 4.61 & 0.142 $\pm$ 0.019 & 0.900$\dagger$ $\pm$ 0.101 \\
NN DGG + ENIR & 80.10 $\pm$ 4.46 & 0.141 $\pm$ 0.019 & 0.899$\dagger$ $\pm$ 0.100 \\
\midrule
BLR & 80.67 $\pm$ 3.34 & 0.144 $\pm$ 0.020 & 0.906 $\pm$ 0.108 \\
Gaussian process & 82.10 $\pm$ 3.40 & 0.135 $\pm$ 0.020 & 0.436$\ddagger$ $\pm$ 0.053 \\
\bottomrule
\addlinespace[\belowrulesep]
\multicolumn{4}{p{0.8\linewidth}}{\footnotesize{Average results of 10-fold cross validation $\pm$ standard deviation. Lower values of MSE and logloss indicate better calibration performance. $^{*}$ Significantly different from Raw, $p < 0.05$. $\dagger$ Significantly different from ENIR full, $p < 0.05$. $\#$ Significantly different from classifier specific calibration, $p < 0.05$. $\ddagger$ Significantly different from RF DGG + ENIR, $p < 0.05$. Significance of the difference determined with Student's paired t-test on 10-fold cross validation results.}}
\end{tabular}
\end{table}

\end{document}